\documentclass[a4paper,fleqn]{cas-sc}

\usepackage[authoryear,longnamesfirst]{natbib}
\usepackage{CJKutf8}
\usepackage[utf8]{inputenc}
\usepackage{pinyin}
\def\tsc#1{\csdef{#1}{\textsc{\lowercase{#1}}\xspace}}
\tsc{WGM}
\tsc{QE}
\tsc{EP}
\tsc{PMS}
\tsc{BEC}
\tsc{DE}

\begin{document}
\begin{CJK}{UTF8}{gbsn}
\let\WriteBookmarks\relax
\def\floatpagepagefraction{1}
\def\textpagefraction{.001}

\shorttitle{A Chinese Spelling Check Framework}

\shortauthors{Nankai Lin et~al.}

\title [mode = title]{A Chinese Spelling Check Framework Based on Reverse Contrastive Learning}                      



%
\author[1]{Nankai Lin}[
                        auid=000,bioid=1,
                        orcid=0000-0003-2838-8273]



\ead{neakail@outlook.com}

\credit{Writing - Original draft preparation, Conceptualization of this study, Methodology}

\affiliation[1]{organization={School of Computer Science and Technology},
    addressline={Guangdong University of Technology}, 
    city={Guangzhou},
    postcode={510000}, 
    country={China}}

\author[2]{Hongyan Wu}

\credit{Methodology, Data curation, Software, Writing- Reviewing and Editing}

\author[2]{Sihui Fu}

\credit{Methodology, Software, Writing- Reviewing and Editing}

\affiliation[2]{organization={School of Information Science and Technology},
    addressline={Guangdong University of Foreign Studies}, 
    city={Guangzhou},
    postcode={510000}, 
    country={China}}

\author%
[2,3]
{Shengyi Jiang}
\cormark[1]
\ead{jiangshengyi@163.com}

\credit{Methodology, Investigation, Writing- Reviewing and Editing}

\author%
[1]
{Aimin Yang}
\cormark[1]
\ead{amyang@gdut.edu.cn}

\credit{Methodology, Investigation, Writing- Reviewing and Editing}

\affiliation[3]{organization={Guangzhou Key Laboratory of Multilingual Intelligent Processing}, 
    city={Guangzhou},
    postcode={510000}, 
    country={China}}

\cortext[cor1]{Corresponding author}



\begin{abstract}
Chinese spelling check is a task to detect and correct spelling mistakes in Chinese text. Existing research aims to enhance the text representation and use multi-source information to improve the detection and correction capabilities of models, but does not pay too much attention to improving their ability to distinguish between confusable words. Contrastive learning, whose aim is to minimize the distance in representation space between similar sample pairs, has recently become a dominant technique in natural language processing. Inspired by contrastive learning, we present a novel framework for Chinese spelling checking, which consists of three modules: language representation, spelling check and reverse contrastive learning. Specifically, we propose a reverse contrastive learning strategy, which explicitly forces the model to minimize the agreement between the similar examples, namely, the phonetically and visually confusable characters. Experimental results show that our framework is model-agnostic and could be combined with existing Chinese spelling check models to yield state-of-the-art performance. 
\end{abstract}






\maketitle

\section{Introduction}

Chinese spelling check (CSC) is an important natural language processing (NLP) task which lays the foundation for many NLP downstream applications, such as optical character recognition (OCR) \citep{wang-etal-2018-hybrid,hong-etal-2019-faspell} or automated essay scoring\citep{uto-etal-2020-neural}. Meanwhile, it is a challenging task which demands the competence comparable to humans, in natural language understanding \citep{liu-etal-2010-visually, liu-hybrid, xin-etal-2014-improved}. Most recent successes on this task are achieved by the non-autoregressive models like BERT, since the length of the output needs to be exactly the same as that of the input, and at the same time each character in the source sequence should share the same position with its counterpart in the target.

As to the classification of Chinese characters, some of them are hieroglyphs while most of them are semantic-phonetic compound characters \citep{norman1988chinese}. Consequently, though it may be impossible to enumerate all spelling mistakes, the error patterns could still be roughly summarized as visual or phonetic errors \citep{chang1995new}. Actually, according to statistics, over 80\% of all spelling mistakes are related to the phonetic resemblance between the characters. If a CSC model could well distinguish between the phonetically and visually confusable characters, it would be of great help to improving its performance in correcting spelling errors. However, currently few methods consider making use of the phonetic and visual information to help tackle the confusable character issue.

On the other hand, self-supervised representation learning has significantly advanced due to the application of contrastive learning \citep{10.5555/3524938.3525087, henaff2020data, oord2018representation, wu2018unsupervised}, whose main idea is to train a model to maximize the agreement between a target example (“anchor”) and a similar (“positive”) example in embedding space, while also maximize the disagreement between this target and other dissimilar (“negative”) examples. \cite{NEURIPS2020_d89a66c7} first allowed contrastive learning to be applied in the fully-supervised setting, namely, Supervised Contrastive Learning (SCL), which could effectively leverage label information to distinguish the positive and negative examples of an anchor. Contrastive learning has brought improvements  to many NLP tasks, including aspect sentiment classification \citep{ke-etal-2021-classic}, text classification \citep{suresh-ong-2021-negatives} and semantic textual similarity \citep{gao-etal-2021-simcse}. With respect to CSC, \cite{li-etal-2022-past} are the first to employ constrastive learning. They proposed an error-driven method to guide the model to learn to tell right and wrong answers.

Enlightened by contrastive learning, in this study we present a simple yet effective CSC framework designed to enhance the performance of existing CSC models. While the objective of contrastive learning is to pull together the similar examples, we creatively propose to pull apart those phonetically or visually similar characters, which would lend a helping hand to the models in distinguishing between confusable characters.

Our contributions could be summarized as follows: 

(1) We entend the idea of contrastive learning to CSC task and propose Reverse Contrastive Learning (RCL) strategy which results in models that better detect and correct spelling errors related to confusable characters. 

(2) We put forward a model-agnostic CSC framework, in which our RCL strategy, as a subsidiary component, could be easily combined with existing CSC models to yield better performance. 

(3) The CSC models equipped with our strategy establish new state-of-the-art on SIGHAN benchmarks.

\section{Related Work}
\subsection{Chinese Spelling Check}
In recent years, the research on CSC has been mainly working on twofold: data generation for CSC and CSC-oriented language models.

\textbf{Data generation for CSC.} \cite{wang-etal-2018-hybrid} proposed a method to automatically construct a CSC corpus, which generates visually or phonologically similar characters based on OCR and ASR recognition technology, respectively, thereby greatly expanding the scale of the corpus. \cite{10.1007/978-3-030-32233-5_37} built two corpora: a visual corpus and phonological one. While the construction of both corpora makes use of ASR technology, the phonological one also uses the conversion between Chinese characters and the sounds of the characters.

\textbf{CSC-oriented language model.}  Recently, researchers have mainly focused on employing language models to capture information in terms of character similarity and phonological similarity, facilitating the CSC task. The models are dominated by neural network-based models \citep{cheng-etal-2020-spellgcn, DBLP:journals/nca/MaHPZX23} , especially pre-trained language models. Related studies have explored the potential semantic modeling capability of pre-trained language models, with BERT being widely utilized as the backbone of CSC models.  However, the methods may lead to overfitting of CSC models to the errors in sentences and disturbance of semantic encoding of sentences, yielding poor generalization \citep{DBLP:journals/corr/abs-2212-04068, DBLP:journals/corr/abs-2305-17721}. To address the issues,\cite{DBLP:journals/corr/abs-2305-03314} proposed the n-gram masking layer to alleviate common label leakage and error disturbance problems. Regarding another line of pre-trained models involving the fusion of textual, visual and phonetic information into pre-trained models,  \cite{wang-etal-2021-dynamic} presented the Dynamic Connected Networks (DCN) based on the non-autoregressive model, which utilizes a Pinyin enhanced candidate generator to generate candidate Chinese characters and then models the dependencies between two neighboring Chinese characters using an attention-based network. SCOPE \citep{DBLP:conf/emnlp/LiWMGY022} enhances the performance of the CSC model by imposing an auxiliary pronunciation prediction task and devising an iterative inference strategy, similar CSC models based on multimodal information as SpellBERT \citep{ji-etal-2021-spellbert}, PLOME \citep{liu-etal-2021-plome} and ReaLiSe \citep{xu-etal-2021-read}. Despite the effectiveness of multimodal information, there are still some inherent problems. Specifically, given that direct integration of phonetic information may affect the raw text representation and weaken the effect of phonetic information, \cite{DBLP:journals/corr/abs-2305-14783} decoupled text and phonetic representation and designed a pinyin-to-character pre-training task to enhance the effect of phonetic knowledge, simultaneously introducing a self-distillation module to prevent the model from overfitting phonetic information. Moreover, to address the issue of characters’ relative positions not being well-aligned in multimodal spaces, \cite{10098240} presented distance fusion and knowledge enhanced framework for CSC to capture relative distances between characters and reconstruct a semantic representation, simultaneously reducing errors caused by unseen fixed expressions.

The aforesaid models introduce character similarity and phonological similarity information from the confusion set, neglecting contextual similarity, which is demonstrated more valuable in the research conducted by \cite{10095675}, where a simple yet effective curriculum learning framework is designed to guide CSC models explicitly to capture the contextual similarity between Chinese characters. 

\subsection{Contrastive Learning for Chinese Spelling Check}
Although contrastive learning has promoted various NLP applications, directly applying it to CSC tasks has limitations. One difficulty lies in constructing suitable examples using data augmentation or existing labels. \cite{DBLP:journals/corr/abs-2210-17168} employed a self-distillation method to construct positive samples for contrastive learning, which uniformed the hidden states of erroneous tokens to be closer to the corresponding correct tokens to learn better feature representation.  LEAD framework \citep{DBLP:conf/emnlp/LiMZLLHLLC022} guides the CSC models to learn better phonetics, vision and definition knowledge from a dictionary exploiting unified contrastive learning, where positive and negative samples are constructed based on the information of character phonetics, glyphs, and definitions from external dictionary. \cite{li-etal-2022-past} proposed the Error-driven Contrastive Probability Optimization (ECOPO) framework. ECOPO optimizes the knowledge representation of the pre-trained model and guides the model to avoid predicting these common features in an error-driven manner. Nevertheless, the method of learning from dictionary leverages the heterogeneous knowledge from a external large-scale dictionary, increasing the cost of training. ECOPO relies on the output of the model to generate negative samples, and hence the quality of these samples is subject to the performance of the model. Unlike the methods, we attempt to adopt the pinyin information from the batch and confusion set to highly efficiently construct the negative samples required for contrastive learning, so as to present accurate contrastive information to the model.

\section{The Chinses spelling check framework}
In this work, we treat spelling check as a non-autoregressive task. Given an input text sequence $X=\{x_1,x_2,…,x_N\}$ and its corresponding Pinyin (Mandarin phonetic transcription) sequence $y=\{y_1,y_2,…,y_N\}$ (where $N$ notates the length of the sequence), the ultimate goal is to automatically detect the incorrect characters in the sequence and then output the rectified target sequence $Z=\{z_1,z_2,…,z_N\}$. The proposed framework for CSC consists of three modules: language representation, spelling correctness and reverse contrastive learning (See Figure 1).

\begin{figure}
  \centering
  \includegraphics[width=1\textwidth]{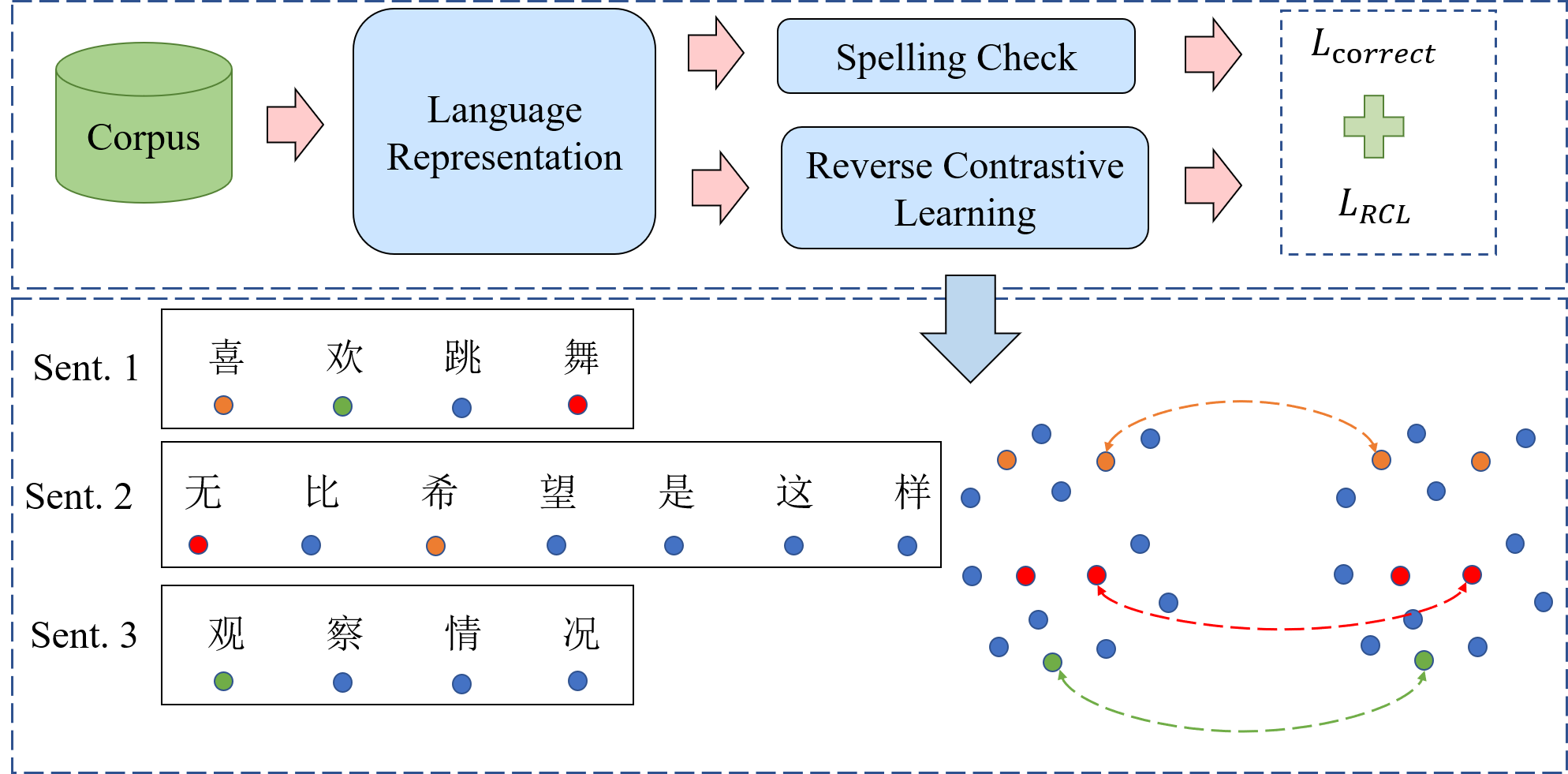}
  \caption{The proposed Chinses spelling check framework.} 
  \label{fig:1} 
\end{figure}

\subsection{Language representation module}
In the language representation module, we adopt a pretrained model to encode the input. Trained on huge unannotated text data, the pretrained model could memorize the language regularity in its parameters and hence offer rich contextual representations to the downstream modules. Given an input sequence $X$, the model would first project the input embeddings to $E_0^t$ and finally yield the language representation $E^t=E_L^t$ after going through multiple hidden encoding layers, among which the output of each layer is denoted as:

\begin{equation}
    E_l^t= Encoder_l (E_{l-1}^t ),l\in[1,L]
\end{equation}
\subsection{Spelling check module}
The spelling check module is a model which performs both spelling check and correctness. More precisely, given the language presentation $E^t$ produced by the previous module, the objective of the model is to detect and rectify those improper characters in the sequence and generate the target sequence $Y$. The training loss of this model is symbolized as $L_{correct}$.
\subsection{Reverse contrastive learning module}
Contrastive learning has been widely used in quite a lot of NLP and machine learning research. The intuition behind it is to train the model in such a way that those similar data points (i.e. samples with the same label or positive samples) will be close to each other, and at the same time far away from dissimilar ones (negative samples).  However, a major obstable in contrastive learning lies in how to construct suitable samples. In particular, different from text classification tasks, it is difficult for CSC to directly use data augmentation or existing labels to construct positive samples. Hence, in this study we propose a reverse contrastive learning (RCL) strategy, in which we only focus on constructing negative samples. Given a Chinses character (a.k.a. the anchor), in our case the negative samples are defined as those characters that sound the same as it (homophones)\footnote{Note that although the Pinyin of a character includes a cluster of letters and a tone, in this work only the letter cluster is taken into consideration, with the tone totally neglected, and thus those characters with the same letter cluster but different tones are still regarded as homophones.}, as well as all those that look similar to it (in its confusion set). Therefore, the learning objective is to maximize the distance between the anchor and its homophones, plus the distance between the anchor and its foes in the confusion set.

Suppose a minibatch which contains $K$ Chinese characters $B_K$ and $I=\{1,…,K \}$ is the character set. Given a character $x_i$ (its Pinyin denoted as $y_i$) and its confusion set $C_i$, all other characters that share the same Pinyin with it in the batch form the set $S=\{s:s\in I,y_s=y_i  \land s \neq i\}$ Then we could define the reverse contrastive loss function based on the Pinyin information:
\begin{small} 
\begin{equation}
    L_{P_i} =  -\frac{1}{|S|}\sum_{s\in S}log\frac{exp(sim(h_i,h_s)/\tau)}{\sum_{k\in I/\{i\} }exp(sim(h_i, h_k))/\tau)}    
\end{equation}
\end{small}
In the meanwhile, all the characters which occur both in the batch $B_K$ and the confusion set $C_i$ constitute the set $W=\{w:w\in C_i \land w \in B_K \}$. Similarly, the loss function based on the confusion sets is:
\begin{small} 
\begin{equation}
    L_{C_i} =  -\frac{1}{|W|}\sum_{w\in W}log\frac{exp(sim(h_i,h_w)/\tau)}{\sum_{k\in I/\{i\} }exp(sim(h_i, h_k))/\tau)}    
\end{equation}
\end{small}
In both equations, $sim(\cdot)$ refers to the cosine similarity function, and $\tau$ is the temperature controlling how sensitive the objective is to similarity. If two characters share a lot in common, a $\tau$ closer to 0 would further emphasize their similarities, while a larger $\tau$ would be more careless of such similarities. The overall reverse contrastive loss is computed across all negative pairs, namely $(x_i,s)$ and $(x_i,w)$, in the batch:
\begin{equation}
    L_{RCL}= \sum_{i=1}^K(L_{P_i}+L_{C_i}) 
\end{equation}

Thus, we obtain the total loss of the whole framework, which is the weighted sum of the reverse contrastive loss and the spelling check training loss mentioned in the previous section:
\begin{equation}
    L= L_{correct}- \alpha L_{RCL}
\end{equation}

Take the three utterances shown in Fig. 1 (a minibatch of size 3) as an example, both (喜, 希) and (舞, 无) are negative pairs as the characters in each pair have the same Pinyin. Besides, (欢, 观) is also a negative pair since ‘观’ occurs in the confusion set of ‘欢’. Our training objective is to maximize the distance between the representations of the characters in each pair, while all the other characters are left alone.

\section{Evaluation protocol}
\textbf{Datasets.} Concerning the training data, we use the Wang271K corpus \footnote{https://github.com/wdimmy/Automatic-Corpus-Generation} \citep{wang-etal-2018-hybrid}, which contains a large amount of automatically generated data. The training sets of SIGHAN 2013 \citep{wu-etal-2013-chinese}, SIGHAN 2014 \citep{yu-etal-2014-overview} and SIGHAN 2015 \citep{tseng-etal-2015-introduction} are also adopted while we evaluate the performance of our proposed strategy on their test sets. Note that different from Wang271k which only include simplified Chinese, the characters in the SIGHAN datasets are traditional Chinese, so we convert all the traditional characters into the simplified ones using OpenCC\footnote{https://github.com/BYVoid/OpenCC} package, just like what previous studies did in their data preprocessing phase. The statistics of the datasets are listed in Table 1. The confusion set we used is from \citep{wang-etal-2018-hybrid}, which is constructed based on visual similarity and phonetic similarity.

It should be pointed out that the annotation quality of SIGHAN13 is not so satisfactory. A prominent issue is that in this benchmark three homophones “的”, “地” and “得”, the frequently used but particually confusable particles, do not get proper annotations, which will probably lead to poor performance even for a good model. To ensure a fair comparison, we follow the approach described in previous studies: removing all the detected and corrected “的”, “地” and “得” from the outputs before reporting the results.

\begin{table}[h]
\centering
\caption{Statistics of the CSC datasets used in this work.}
\begin{tabular}{p{2.3cm}p{0.95cm}p{1.9cm}p{0.95cm}}
\hline
Train Set & \#Sent & Avg. Length & \#Errors \\
\hline
Wang271K & 271329 & 44.4 & 271329 \\
SIGHAN 2013 & 700 & 49.2 & 350 \\
SIGHAN 2014 & 3435 & 49.7 & 3432 \\
SIGHAN 2015 & 2339 & 30.0 & 2339 \\
\hline
Test Set & \#Sent & Avg. Length & \#Errors \\
\hline
SIGHAN 2013 & 1000 & 74.1 & 996 \\
SIGHAN 2014 & 1062 & 50.1 & 529 \\
SIGHAN 2015 & 1100 & 30.5 & 550 \\
\hline
\end{tabular}
\label{tab:accents}
\end{table}

\textbf{Setups.} To demonstrate that our CSC framework is not subject to some specific model, we try to implement it using both DCN and ReaLiSe. For DCN, RoBERTa \citep{10.1109/TASLP.2021.3124365} is employed as the pre-trained model in the language representation module, while the spelling check module is comprised of dynamic connected scorer and Pinyin-enhanced candidate generator. The whole network is optimized using AdamW with learning rate of 5e-5 and trained at batch size 32 for 20 epochs. When it comes to ReaLiSe, we take BERT \citep{devlin-etal-2019-bert} as the pre-trained model. We adopt the same optimizer as described above and train the model at batch size 32 for 10 epochs. Furthermore, we use warmup and linear decay for learning rate scheduling. As for the hyperparameters of the models themselves, we follow the settings in the papers introducing these two models. In addition, to calculate $L_{correct}$, we choose cross entropy. As for the weight of the reverse contrastive loss, we adopt parameter search to find the final value.

\begin{table*}[!h]
\centering
\caption{SIGHAN 2013 Test Results. $\ast$ means the result is post-processed by \cite{xu-etal-2021-read}}
\begin{tabular}{ccccccc}
\hline
Model & D-P & D-R & D-F & C-P & C-R & C-F \\
\hline
{\makecell[c]{Sequence Labeling \\ \cite{wang-etal-2018-hybrid}}} & 54.0 & 69.3 & 60.7 & - & - & 52.1 \\
{\makecell[c]{FASPell \\ \cite{hong-etal-2019-faspell}}} & 76.2 & 63.2 & 69.1 & 73.1 & 60.5 & 66.2 \\
{\makecell[c]{BERT \\ \cite{devlin-etal-2019-bert}}} & 79.0 & 72.8 & 75.8 & 77.7 & 71.6 & 74.6 \\
BERT$\ast$  & 85.0 & 77.0 & 80.8 & 83.0 & 75.2 & 78.9 \\
{\makecell[c]{SpellGCN \\ \cite{cheng-etal-2020-spellgcn}}} & 80.1 & 74.4 & 77.2 & 78.3 & 72.7 & 75.4 \\
SpellGCN$\ast$  & 85.7 & 78.8 & 82.1 & 84.6 & 77.8 & 81.0 \\
\hline
DCN & 84.7 & 77.2 & 80.8 & 83.5 & 76.1 & 79.6 \\
DCN-based RCL Framework & 
{\makecell[c]{\textbf{86.3} \\ \textbf{(+1.6)}}} & {\makecell[c]{\textbf{78.2} \\ \textbf{(+1.0)}}} & 
{\makecell[c]{\textbf{82.0} \\ \textbf{(+1.2)}}} & 
{\makecell[c]{\textbf{85.3} \\ \textbf{(+1.8)}}} & 
{\makecell[c]{\textbf{77.3} \\ \textbf{(+1.2)}}} & 
{\makecell[c]{\textbf{81.1} \\ \textbf{(+1.5)}}} \\
\hline
ReaLiSe & 86.8 & 80.5 & 83.6 & 85.6 & 79.4 & 82.4 \\
ReaLiSe-based RCL Framework & 
{\makecell[c]{\textbf{87.8} \\ \textbf{(+1.0)}}} & 
{\makecell[c]{\textbf{81.2} \\ \textbf{(+0.7)}}} & 
{\makecell[c]{\textbf{84.3} \\ \textbf{(+0.7)}}} & 
{\makecell[c]{\textbf{86.5} \\ \textbf{(+0.9)}}} & 
{\makecell[c]{\textbf{80.0} \\ \textbf{(+0.6)}}} & 
{\makecell[c]{\textbf{83.2} \\ \textbf{(+0.8)}}} \\
\hline
\end{tabular}
\label{tab:accents}
\end{table*}

\begin{table*}[!h]
\centering
\caption{SIGHAN 2014 Test Results.}
\begin{tabular}{ccccccc}
\hline
Model & D-P & D-R & D-F & C-P & C-R & C-F \\
\hline
Sequence Labeling & 51.9 & 66.2 & 58.2 & - & - & 56.1 \\
FASPell & 61.0 & 53.5 & 57.0 & 59.4 & 52.0 & 55.4 \\
BERT & 64.5 & 68.6 & 66.5 & 62.4 & 66.3 & 64.3 \\
SpellGCN & 65.1 & 69.5 & 67.2 & 63.1 & 67.2 & 65.3 \\
\hline
DCN & 62.9 & 65.9 & 64.4 & 61.3 & 64.2 & 62.7 \\
Our Framework(DCN) & 
{\makecell[c]{\textbf{66.7} \\ \textbf{(+3.8)}}} & 
{\makecell[c]{\textbf{67.6} \\ \textbf{(+1.7)}}} & 
{\makecell[c]{\textbf{67.2} \\ \textbf{(+2.8)}}} & 
{\makecell[c]{\textbf{65.2} \\ \textbf{(+3.9)}}} & 
{\makecell[c]{\textbf{66.1} \\ \textbf{(+1.9)}}} & 
{\makecell[c]{\textbf{65.7} \\ \textbf{(+3.0)}}} \\
\hline
ReaLiSe & 64.6 & 67.1 & 65.9 & 62.8 & 65.2 & 64.0 \\
Our Framework(ReaLiSe) & 
{\makecell[c]{\textbf{65.8} \\ \textbf{(+1.2)}}} & 
{\makecell[c]{\textbf{69.2} \\ \textbf{(+2.1)}}} & 
{\makecell[c]{\textbf{67.5} \\ \textbf{(+1.6)}}} & 
{\makecell[c]{\textbf{64.5} \\ \textbf{(+1.7)}}} & 
{\makecell[c]{\textbf{67.9} \\ \textbf{(+2.7)}}} & 
{\makecell[c]{\textbf{66.2} \\ \textbf{(+2.2)}}} \\
\hline
\end{tabular}
\label{tab:accents}
\end{table*}

\section{Experiments}
Table 2, 3 and 4 report the results of our RCL framework comparing to other models on SIGHAN 2013, SIGHAN 2014 and SIGHAN 2015 benchmarks respectively. Since our method is experimented in the fine-tuning stage, the data used is limited to four data such as SIGHAN 2013, we only compared the models that are also optimized in the fine-tuning stage. We observe that our reverse constrastive strategy produces consistent improvements across different datasets. From Table 2, we see that with DCN, the F-values of our spelling check framework increase 1.2\% on error detection and 1.5\% on correction on SIGHAN 2013, compared to the vanilla DCN. Moreover, the ReaLiSe-based framework achieves the highest scores, with 84.3\% F-value in terms of detection and 83.2\% correction.

On SIGHAN 2014, our method also achieves significant improvenments. The DCN-based framework outperforms the vanilla model by 2.8\% (detection) and 3.0\% (correction) regarding the F score, while the ReaLiSe-based framework by 1.6\% and 2.2\%. Similarly, on SIGHAN 2015, the DCN equipped with our reverse constrative strategy obtains a higher F1 score than all the other state-of-the-art models. All the results show the effectiveness of our proposed method, and well demonstrate that it is not model-dependent.
\begin{table*}[h]
\centering
\caption{SIGHAN 2015 Test Results.}
\begin{tabular}{ccccccc}
\hline
Model & D-P & D-R & D-F & C-P & C-R & C-F \\
\hline
Sequence Labeling & 56.6 & 69.4 & 62.3 & - & - & 57.1 \\
FASPell & 67.6 & 60.0 & 63.5 & 66.6 & 59.1 & 62.6 \\
Soft-Masked BERT & 73.7 & 73.2 & 73.5 & 66.7 & 66.2 & 66.4 \\
BERT & 74.2 & 78.0 & 76.1 & 71.6 & 75.3 & 73.4 \\
SpellGCN & 74.8 & 80.7 & 77.7 & 72.1 & 77.7 & 75.9 \\
\hline
ReaLiSe & 75.3 & 79.3 & 77.2 & 72.8 & 76.7 & 74.7 \\
Our Framework(ReaLiSe) & 
{\makecell[c]{\textbf{78.0} \\ \textbf{(+2.7)}}} & 
{\makecell[c]{\textbf{81.3} \\ \textbf{(+2.0)}}} & 
{\makecell[c]{\textbf{79.6} \\ \textbf{(+2.4)}}} & 
{\makecell[c]{\textbf{76.4} \\ \textbf{(+3.6)}}} & 
{\makecell[c]{\textbf{79.7} \\ \textbf{(+3.0)}}} & 
{\makecell[c]{\textbf{78.0} \\ \textbf{(+3.3)}}} \\
\hline
DCN & 76.5 & 80.4 & 78.4 & 74.4 & 78.2 & 76.2 \\
Our Framework(DCN) & 
{\makecell[c]{\textbf{78.4} \\ \textbf{(+1.9)}}} & 
{\makecell[c]{\textbf{81.3} \\ \textbf{(+0.9)}}} & 
{\makecell[c]{\textbf{79.8} \\ \textbf{(+1.4)}}} & 
{\makecell[c]{\textbf{76.3} \\ \textbf{(+1.9)}}} & 
{\makecell[c]{\textbf{79.1} \\ \textbf{(+0.9)}}} & 
{\makecell[c]{\textbf{77.7} \\ \textbf{(+1.5)}}} \\
\hline
\end{tabular}
\label{tab:accents}
\end{table*}

\begin{table*}[!h]
\centering
\caption{Ablation Study Results.}
\begin{tabular}{ccccccc}
\hline
Model & D-P & D-R & D-F & C-P & C-R & C-F \\
\hline
Our Framework(ReaLiSe) & 78.0 & 81.3 & 79.6 & 76.4 & 79.7 & 78.0 \\
\hline
w/o confusion set & 
{\makecell[c]{77.0 \\ (-1.0)}} & 
{\makecell[c]{80.6 \\ (-0.7)}} & 
{\makecell[c]{78.8 \\ (-0.8)}} & 
{\makecell[c]{75.6 \\ (-0.8)}} & 
{\makecell[c]{79.1 \\ (-0.6)}} & 
{\makecell[c]{77.3 \\ (-0.7)}} \\
\hline
w/o pinyin & 
{\makecell[c]{76.3 \\ (-1.7)}} & 
{\makecell[c]{80.2 \\ (-1.1)}} & 
{\makecell[c]{78.2 \\ (-1.4)}} & 
{\makecell[c]{74.5 \\ (-1.9)}} & 
{\makecell[c]{78.4 \\ (-1.3)}} & 
{\makecell[c]{76.4 \\ (-1.6)}} \\
\hline
\end{tabular}
\label{tab:accents}
\end{table*}

We conduct ablation studies on different modules, taking the results of the Realise-based framework on the SIGHAN 2015 dataset as an example, and the experimental results are shown in Table 5. It can be seen that when the pinyin-based RCL module is removed, the detection-level F-value and the correction-level F-value drop by 1.4\% and 1.6\% respectively, while the confusion-set-based RCL module is removed, the detection-level F-value and the correction-level F value only drop by 0.8\% and 0.7\%. It can be seen that pinyin information plays a greater role in improving the performance of the model.

To better understand the importance of our RCL strategy, we investigate the impacts of different weight settings for $L_{RCL}$ (i.e. the α in Equation 5) on the performance of our framework. Since this strategy plays a subsidiary role to the spelling check module, $L_{RCL}$ should not account for a large proportion in the total loss, suggesting a small $\alpha$. Table 6 presents the experimental results on SIGHAN 2013 benchmark using the ReaLiSe-based model, with $\alpha$ ranging from 0.0001 to 0.1. We could see that when $\alpha$=0.1, RCL suffers a remarkable reduction of 4.8\% and 1.8\% on both detection and correction evaluations, compared to its vanilla counterpart. We believe the reduction is due to the model’s over-emphasis on distinguishing homophones other than learning to conduct spelling check and correction. If we keep decreasing the value of $\alpha$ till 0.001, RCL starts to be on the same level with the original model. In other words, the knowledge acquired from confusable characters probably kicks in when $\alpha$ is less than 0.001.

To determine the optimal $\alpha$ used in the final report, we further explore its different settings within 0.001 for both ReaLiSe-based and DCN-based framworks. As illustrated in Figure 2, on SIGHAN 2015, the optimal $\alpha$ for the ReaLiSe-based framework is 0.0001, different from the one 0.0007 on SIGHAN 2013 and SIGHAN 2014. As for the DCN-based framework, $\alpha$=0.0005 is our best choice.
\begin{table}[h]
\centering
\begin{tabular}{p{0.75cm}p{0.65cm}p{0.7cm}p{0.65cm}p{0.65cm}p{0.65cm}p{0.65cm}}
\hline
Weight & D-P & D-R & D-F & C-P & C-R & C-F \\
\hline
- & 86.8 & 80.5 & 83.6 & 85.6 & 79.4 & 82.4 \\
0.1 & 82.3 & 75.6 & 78.8 & 84.2 & 77.3 & 80.6 \\
0.01 & 86.6 & 79.2 & 82.7 & 85.5 & 78.2 & 81.7 \\
0.001 & 86.9 & 80.4 & 83.5 & 85.7 & 79.3 & 82.4 \\
0.0001 & 87.6 & 80.5 & 83.9 & 86.6 & 79.6 & 82.9 \\
\hline
\end{tabular}
\caption{Performance of RCL framework (ReaLiSe-based) with different weights of $L_{RCL}$ on SIGHAN 13}
\label{tab:accents}
\end{table}

\begin{figure}[!h]
  \centering
  \includegraphics[width=0.8\textwidth]{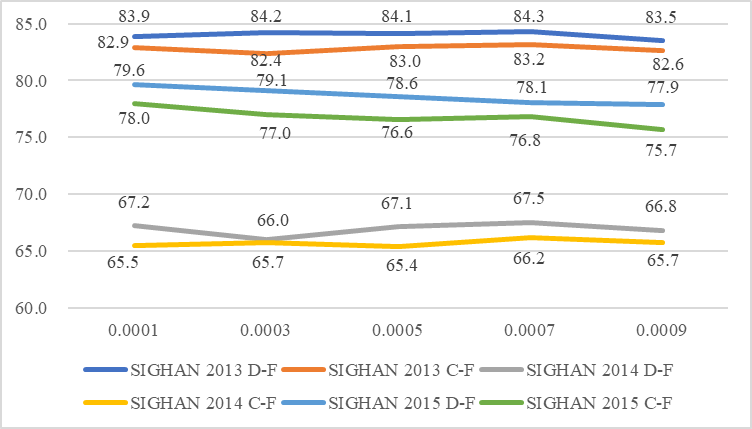}
  \caption{Parameter Exploration Experimental Results of RCL framework (ReaLiSe-based).} 
  \label{fig:2} 
\end{figure}

\begin{figure}[!h]
  \centering
  \includegraphics[width=0.8\textwidth]{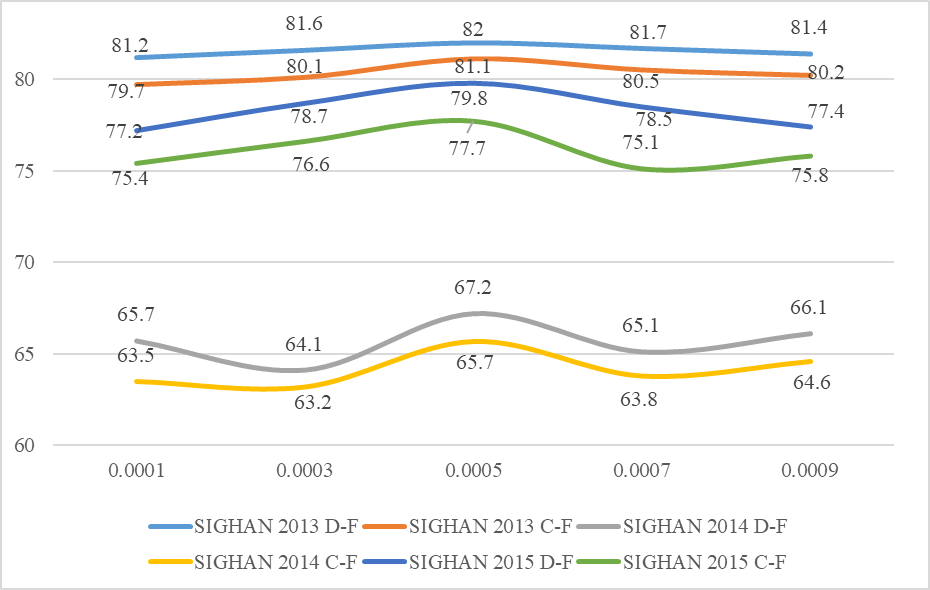}
  \caption{Parameter Exploration Experimental Results of RCL framework (DCN-based).} 
  \label{fig:3} 
\end{figure}

\section{Case study}
As shown in Table 7, we sample predictions from vanilla DCN and DCN-based RCL framework to examine the effect of reverse contrastive learning on spelling check. We see that the DCN-based RCL framework could indeed recognize and correct those confusable characters that vanilla DCN could not. In Sample 1, DCN-based RCL manages to detect all the two spelling errors in the input, while vanilla DCN can only detect one. Note that the output “师” of DCN can form a correct word with “教”, but it somehow causes semantic incoherence for the whole sentence. Similarly, in Sample 2 “教” could co-exist with “育” to produce a reasonable word, but it is not the best choice in terms of semantics. The third example is caused by the misuse of “希望” and “习惯”. Both models output the gold character “希” as a correction for the input homophone ‘习’, but only DCN-based RCL also yields the correct “望” for “惯”, one of the characters in its confusion set. These cases demonstrate that our RCL strategy does equip the vanilla CSC model with the ability to find those phonetically and visually confusable characters.

\begin{table*}[!h]
\centering
\caption{Prediction samples from vanilla DCN and DCN-based RCL framework}
\begin{tabular}{p{3.5cm}|p{11.5cm}}
\hline
\multicolumn{2}{l}{Sample 1} \\
\hline
\multirow{2}{*}{Input} & 我带来一个地图可是我看不动\emph{(\dong4)}，也不知道教师\emph{(\shi1)}在哪里。\\
& I brought a map but I cannot move, and I don’t know where the teacher is. \\
\hline
\multirow{2}{*}{Ground truth} & 我带来一个地图可是我看不懂\emph{(\dong3)}，也不知道教室\emph{(\shi4)}在哪里。\\
& I brought a map but I cannot understand, and I don’t know where the classroom is. \\
\hline
\multirow{2}{*}{\makecell[c]{Prediction from DCN}} & 我带来一个地图可是我看不懂\emph{(\dong3)}，也不知道教师\emph{(\shi1)}在哪里。\\
& I brought a map but I cannot understand, and I don’t know where the teacher is. \\
\hline
\multirow{2}{*}{\makecell[l]{Prediction from DCN \\ -based RCL framework}} & 我带来一个地图可是我看不懂\emph{(\dong3)}，也不知道教室\emph{(\shi4)}在哪里。\\
& I brought a map but I cannot move, and I don’t know where the teacher is. \\
\hline
\multicolumn{2}{l}{Sample 2} \\
\hline
\multirow{2}{*}{Input} & 而且老师也很难教师\emph{(\shi1)}。\\
& And teachers are hard to teacher. \\
\hline
\multirow{2}{*}{Ground truth} & 而且老师也很难教书\emph{(\shu1)}。\\
& And teachers are hard to teach. \\
\hline
\multirow{2}{*}{\makecell[c]{Prediction from DCN}} & 而且老师也很难教育\emph{(\yu4)}。\\
& And teachers are hard to educate. \\
\hline
\multirow{2}{*}{\makecell[l]{Prediction from DCN \\ -based RCL framework}} & 而且老师也很难教书\emph{(\shu1)}。\\
& And teachers are hard to teach. \\

\hline
\multicolumn{2}{l}{Sample 3} \\
\hline
\multirow{2}{*}{Input} & 我等你的答案，习\emph{(\xi3)}惯\emph{(\guan4)}你下个月也有时间。\\
& I am waiting for you answer, and am used to you will have time next month. \\
\hline
\multirow{2}{*}{Ground truth} & 我等你的答案，希\emph{(\xi1)}望\emph{(\wang4)}你下个月也有时间。\\
& I am waiting for you answer, and hope you will have time next month. \\
\hline
\multirow{2}{*}{\makecell[c]{Prediction from DCN}} & 我等你的答案，希\emph{(\xi1)}惯\emph{(\guan4)}你下个月也有时间。\\
& I am waiting for you answer, and hope and am used to you will have time next month. \\
\hline
\multirow{2}{*}{\makecell[l]{Prediction from DCN \\ -based RCL framework}} & 我等你的答案，希\emph{(\xi1)}望\emph{(\wang4)}你下个月也有时间。\\
& I am waiting for you answer, and hope you will have time next month. \\
\hline
\end{tabular}
\label{tab:accents}
\end{table*}


\section{Conclusion}
In this article, we introduce a novel framework for CSC, one of whose core components is reverse constrastive learning, a simple yet effective strategy inspired by contrastive learning. Our method takes advantage of the information provided by Pinyin and confusion sets to construct negative examples and aims to maximize the disagreement between the anchor and these examples. Experimental results demonstrate that this model-indenpent strategy could assist existing CSC models in better recognizing phonetically and visually confusable characters and making corrections. Next we will try to integrate other sorts of confusable characters to further enhance our framework.

\printcredits

\bibliographystyle{cas-model2-names}

\bibliography{cas-refs}

\end{CJK}
\end{document}